\documentclass{article}

\usepackage[preprint]{neurips_2025}

\usepackage[utf8]{inputenc} 
\usepackage[T1]{fontenc}    
\usepackage{hyperref}       
\usepackage{url}            
\usepackage{booktabs}       
\usepackage{amsfonts}       
\usepackage{nicefrac}       
\usepackage{microtype}      
\usepackage{xcolor}         
\usepackage{amsmath}
\usepackage{appendix}
\usepackage{amsthm}
\usepackage{algorithm}
\usepackage{algpseudocode}
\usepackage{graphicx} 
\usepackage{wrapfig}

\title{Hierarchical Stochastic Differential Equation Models for Latent Manifold Learning in Neural Time Series}

\author{
  Pedram Rajaei \\
  Department of Biomedical Engineering \\
  University of Houston \\
  Houston, TX, USA \\
  \texttt{prajaei@cougarnet.uh.edu} \\
  \And
  Maryam Ostadsharif Memar \\
  Department of Electrical and Computer Engineering \\
  Isfahan University of Technology \\
  Isfahan, Iran \\
  \texttt{m.ostadsharif@ec.iut.ac.ir} \\
  \And
  Navid Ziaei \\
  Department of Electrical and Computer Engineering \\
  Isfahan University of Technology \\
  Isfahan, Iran \\
  \texttt{navid.ziaei94@gmail.com} \\
  \And
  Behzad Nazari \\
  Department of Electrical and Computer Engineering \\
  Isfahan University of Technology \\
  Isfahan, Iran \\
  \texttt{nazari@iut.ac.ir} \\
  \And
  Ali Yousefi \\
  Department of Biomedical Engineering \\
  University of Houston \\
  Houston, TX, USA \\
  \texttt{aliyousefi@uh.edu} \\
}

\begin{document}

\maketitle

\begin{abstract}
The manifold hypothesis suggests that high-dimensional neural time series lie on a low-dimensional manifold shaped by simpler underlying dynamics. To uncover this structure, latent dynamical variable models such as state-space models, recurrent neural networks, neural ordinary differential equations, and Gaussian Process Latent Variable Models are widely used. We propose a novel hierarchical stochastic differential equation (SDE) model that balances computational efficiency and interpretability addressing key limitations of existing methods. Our model assumes the trajectory of a manifold can be reconstructed from a sparse set of samples from the manifold trajectory. The latent space is modeled using Brownian bridge SDEs, with points — specified in both time and value — sampled from a multivariate marked point process. These Brownian bridges define the drift of a second set of SDEs, which are then mapped to the observed data. This yields a continuous, differentiable latent process capable of modeling arbitrarily complex time series as the number of manifold points increases. We derive training and inference procedures and show that the computational cost of inference scales linearly with the length of the observation data. We then validate our model on both synthetic data and neural recordings to demonstrate that it accurately recovers the underlying manifold structure and scales effectively with data dimensionality.

\end{abstract}

\section{Introduction}
The manifold hypothesis proposes that many real-world high-dimensional time series data lie on a lower-dimensional latent manifold embedded within the high-dimensional space \citep{ref18}. For example, in neural data, there is evidence that neural population activity in the auditory cortex evolves within a low-dimensional space, governed by a stimulus-dependent initial condition \citep{ref19}. Similarly, in speech data, the signal resides on a low-dimensional manifold within the higher-dimensional ambient space of sound pressure time series, constrained by the physical properties of the human vocal tract and the phonetic rules of a language \citep{ref20}. Numerous modeling techniques have been developed to characterize and infer such latent manifolds. These include, but are not limited to, dynamical principal component analysis \citep{ref21}, State Space Models (SSM)\citep{ref1},dynamical autoencoders \citep{ref22}, switching state-space models \citep{ref23}, nonparametric methods such as Gaussian and Dirichlet processes \citep{ref24, ref5, ref4}, and locality-preserving techniques such as t-SNE and UMAP \citep{ref30, ref31}.

In this paper, we focus on SSMs  and, in particular, expanding this framework for analysis of high-dimensional time series data that arises in neuroscience experiments. SSM consists of a state equation, which characterizes the dynamics of the system, and an observation equation that describes how the state dynamics is observed or measured. Recent advancement in the SSM domain has moved towards models with flexible mappings for the state and/or observation equations, such as Deep SSM (DSSM) \citep{ref2}, Deep Kalman Filter (DKF) \citep{ref3}, Gaussian Process Dynamical Models (GPDM) \citep{ref4}, and models with a nonparametric characterization of either of the state or observation equations, such as those explained by Gaussian Process State Space Models (GPSSM) \citep{ref5} and the Latent Discriminative Generative Decoder (LDGD) \citep{ref6}. These realizations of SSM are useful for the discovery of the latent manifolds, and they can potentially be used in explaining behavior or underlying cognitive processes embedded in neural data \citep{ref4}. Despite utilities of these models in characterizing time-series data, they have their own limitations. Models such as DKF and GPSSM offer high predictive performance but often struggle to capture rhythmic dynamics present in the data; thus they fail to reflect the underlying neural mechanisms. Additionally, these models are data greedy, which poses a challenge given the limited availability of data in neuroscience.  DSSMs, which combine Recurrent Neural Networks (RNNs) for latent dynamics with DNN observation models, are increasingly used for neural data. However, they often require additional structural constraints to reliably capture rhythmic or oscillatory patterns \citep{ref54}. Beyond SSMs, RNNs and Long Short-Term Memory (LSTM) networks can model nonlinear, long-term dependencies in time series data \citep{ref11}, but their internal representations are difficult to interpret. Moreover, training these models can be unstable due to issues such as vanishing gradients and overfitting, particularly in high-dimensional settings \citep{ref10}.

\begin{figure}[t]
    \centering
    \includegraphics[width=0.9\linewidth]{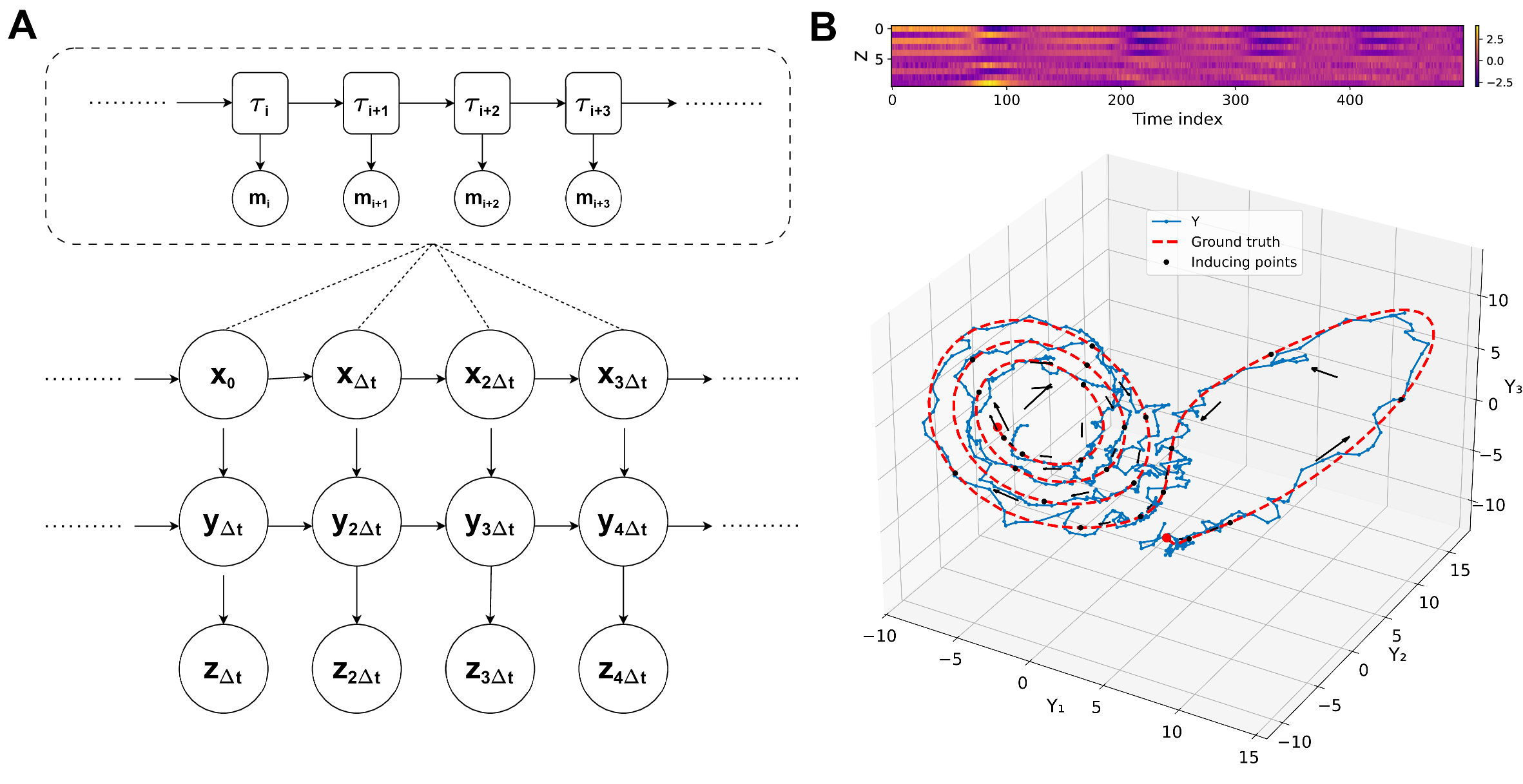}
    \caption{\textbf{Hierarchical SDE Model Structure and its Application in Decoding Lorenz System Trajectories:} (A) Shows the graphical representation of our proposed model, where inducing points ($\tau_i$,$m_i$) go through two cascade SDEs ($X_t$,$Y_t$) followed by projection to the observation domain ($Z_k$). (B)Top row shows the observed data used as input to our proposed model. This data is generated by projecting the Lorenz trajectory through a 10-dimensional linear mapping followed by additive noise. The bottom row shows the simulated Lorenz trajectory and its estimation by our model, along with the timing of the inducing points. Our model robustly infers the nonlinear and complex dynamics of the Lorenz attractor, with the inducing point timings efficiently adapted to capture both fast and slow transitions in the trajectory.}
    \label{fig:lorenz}
\end{figure}

In this research, we focus on a class of SSMs that balance interpretability with a high level of expressive power. A key premise of our approach is that capturing the underlying manifold dynamics requires only a sparse set of trajectory samples over time. This assumption is supported by findings in neuroscience, where latent neural manifolds often exhibit smooth, low-dimensional dynamics as it is observed in neural data recorded during motor task and also cognitive processes \citep{ref34, ref55}. To model high-dimensional time-series data through low-dimensional latent representations, we introduce a hierarchical stochastic modeling framework which propagate samples of manifold trajectories through model layers to generate observed data. Our proposed model incorporates two layers of stochastic differential equations (SDEs) to construct the latent space and then map them to observed data. The first layer consists of a set of SDEs, each defined by a Brownian bridge process \citep{ref16}, where the time and values of the bridge points — i.e., inducing points — are derived from a multivariate marked point process \citep{ref15}. The trajectories of these processes define the drift term for a second set of SDEs. The cascaded structure of these SDEs, along with the sequence of inducing points, shapes the evolution of the latent processes. In our model, the observed data at discrete time points are modeled as a noisy linear projection of the latent process, which facilitates capturing its underlying dynamics. We prove that the latent process can serve as a universal approximator for time-series data in the limit of a large number of inducing points. We then derive the inference and training procedure for the framework, enabling identification of the inducing points and the underlying manifold, and their projection onto the data. We further evaluate the computational efficiency of our proposed model and show that it outperforms Gaussian Process (GP)-based models. Figure~\ref{fig:lorenz} shows the graphical model of the proposed framework along with a conceptual illustration based on a simulated trajectory of the Lorenz system.

The remainder of this paper is organized as follows. In Section 2, we present the details of our hierarchical SDE model, discuss its properties, and derive the associated inference and training pipeline. Section 3 focuses on the application of our model to both simulated data and neural recordings. For the neural recordings, we analyze samples of a monkey’s cortical data collected during an arm-reaching task. Sections 4 and 5 explore various aspects of the proposed framework, and we conclude by highlighting its potential for characterizing time-series data.

\section{Materials \& Methods}
\label{secc}
In this section, we first define the components of our model, including the generation of inducing points and the SDEs that map these points to the observed time series data. We establish the universal approximation properties of the proposed model and then develop its training and inference procedures. 

\subsection{Hierarchical SDE Framework}
\label{hsde}
Figure~\ref{fig:lorenz}A illustrates the model structure. The inducing points are a set of event-value pairs that sample the underlying manifold in both time and value space. Each event is characterized by a timestamp $t_i$ and an associated mark vector $\vec{m}_i$. The model considers an arbitrary finite sequence of these pairs, represented as the set $\mathcal{I} = \{ (\vec{m}_i, t_i);\ i = 1, 2, \dots \}$.
The joint probability distribution over any $L$ of these pairs is given by:

\begin{equation}
\label{eq1}
    p\left( \{(t_i, \vec{m}_i)\}_{i=1}^L \right) = \prod_{i=1}^L \lambda(t_i \mid \mathcal{H}_i)\, p(\vec{m}_i \mid t_i, \mathcal{H}_i)
\end{equation}

where \( \lambda(t_i \mid \mathcal{H}_i) \) models the event occurrence rate conditioned on the history of previous events \( \mathcal{H}_i \), and \( p(\vec{m}_i \mid t_i, \mathcal{H}_i) \) defines the mark distribution conditioned on the event time \( t_i \) and the history \( \mathcal{H}_i \) \citep{ref44}. The sequence of event times can also be described using waiting times, defined as $\tau_i = t_i - t_{i-1}$, which transforms the process into a renewal marked point process \citep{ref15}. Under this formulation, the joint distribution of event-value pairs is defined by:
\begin{equation}
\label{eq2}
    p\left( \{(t_i, \vec{m}_i)\}_{i=1}^L \right) = \prod_{i=1}^L P(\tau_i \mid \mathcal{H}_i) \cdot p(\vec{m}_i \mid \tau_i, \mathcal{H}_i) \approx \prod_{i=1}^L p(\tau_i)\, p(\vec{m}_i)
\end{equation}

In the specific case of this joint distribution as shown in the right end of the above equation, we assume that the waiting times \( \tau_i \) are independent of the previous events. Similarly, we assume that \( \vec{m}_i \) is independent of the previous events and current waiting time. The choice of waiting time distribution and conditional independence facilitates the training and inference for our proposed model, as it will be discussed in Section \ref{inference}. We model the waiting times using a Gamma distribution, while the values \( \vec{m}_i \) are assumed to follow a multivariate normal distribution. The choice of the Gamma distribution for modeling waiting times, along with the prior for the event-value distribution, is explained in the Appendix~\ref{app:priors}.

With the inducing points generated, we now introduce the remaining components of the model that map these points to the observed time-series data. Let \( Z_k \in \mathbb{R}^M \) denote the observed data at discrete time points \( k = 1, \dots, K \), which are modeled as functions of an underlying continuous latent process \( Y_t \in \mathbb{R}^D \), where \( D \ll M \). \( Y_t \) evolves according to another latent process \( X_t \), which has the same dimension as \( Y_t \). The latent process \( X_t \) is modeled using a set of SDEs, where the process is constrained to reach the mark values at times specified by the inducing time points - i.e, a Brownian bridge SDE \citep{ref45}. Evolution of state process in each dimension \( d = 1, \dots, D \) is defined by:

\begin{equation}
    d x_t^d = \mu_t^d \, dt + \sigma_t^d \, dw_t^d,
\end{equation}

where \( w_t^d \) is a standard Wiener process, \( \mu_t^d \) is the drift term, and \( \sigma_t^d \) is the time-dependent diffusion coefficient. For \( t \in [t_i, t_{i+1}) \), which corresponds to the waiting time \( \tau_{i+1} = t_{i+1} - t_i \), the drift and diffusion terms are defined as:

\begin{equation}
    \mu_t^d = \frac{m_{i+1}^d - x_t^d}{t_{i+1} - t}, \quad \sigma_t^d = \sqrt{\frac{(t_{i+1} - t)(t - t_i)}{t_{i+1} - t_i}},
\end{equation}

where \( m_{i+1}^d \) is the \( d \)-th component of the mark vector \( \vec{m}_{i+1} \), setting the value that the process must reach at the event time \( t_{i+1} \). The latent process \( Y_t \) evolves according to:
\begin{equation}
    d y_t^d = x_t^d \, dt + \sigma_y^d \, d\nu_t^d,
\end{equation}
where the drift term for \( Y_t \) is defined by \( X_t \), \( \nu_t^d \) is a standard Wiener process, and \( \sigma_y^d \) is the diffusion coefficient for the \( d \)-th component. Finally, the observations \( Z_k \) are defined as:

\begin{equation}
\label{eq6}
    Z_k = W
    \begin{pmatrix}
        y_{k\Delta t}^1 \\
        \vdots \\
        y_{k\Delta t}^D
    \end{pmatrix}
    + \varepsilon_k,
\end{equation}

where \( \Delta t \) denotes the sampling interval at which the observations \( Z_k \) are collected, \( W \in \mathbb{R}^{M \times D} \) is a linear projection matrix, and \( \varepsilon_k \sim \mathcal{N}(0, R) \) represents Gaussian noise with covariance \( R \in \mathbb{R}^{M \times M} \).

\subsection{Model Properties}
In the next section, we discuss the key attributes of our proposed model, including its universal approximation capability, nonparametric nature, and computational cost. We also describe strategies for managing the growth of inducing points.

\subsubsection{Universal Approximation Property}
When the process is deterministic and the inducing points are equally spaced, we can rely on the sampling theorem which suggests that a signal can be completely reconstructed from its samples \citep{ref33}. In simple terms, any continuous function can be reconstructed from properly sampled data points. Here, we extend a similar idea to hierarchical SDEs using the inducing points.

\textbf{Theorem:}  
Let \( f \in C([0, T]) \) be a continuous function and let \( \varepsilon > 0 \) be arbitrary value. Then there exists a choice of inducing points such that the expected process \( \mathbb{E}[Y_t] \) uniformly approximates the integral
\begin{equation}
    g(t) = \int_0^t f(s) \, ds
\end{equation}

within error \( \varepsilon \), in the sense that
\begin{equation}
    \sup_{t \in [0, T]} \left| \mathbb{E}[Y_t] - g(t) \right| < \varepsilon
\end{equation}

\textbf{Proof:} With our model, we approximate a continuous function \( g(t) \) on \( [0, T] \) using the process \( Y_t = \int_0^t X_s \, ds + \sigma W_t \), where \( X_t \) is a Brownian bridge and $\sigma$ is a noise variance. Let’s assume we sample \( X_t \) at \( N \) uniformly spaced inducing points \( \{t_i\} \) and assign marks \( m_i \) to each time. We define \( X_t^N \) using piecewise linear interpolation based on these inducing points. Since such functions are dense in \( C([0, T]) \), we can choose \( m_i \) so that \( X_t^N \to f(t) \) uniformly for any continuous \( f(t) \). We also define \( Y_t^N = \int_0^t X_s^N \, ds \). Then, \( \mathbb{E}[Y_t^N] \to g(t) = \int_0^t f(s) \, ds \) uniformly, as integration preserves uniform convergence. The noise term \( \sigma W_t \) has zero mean and variance \( \sigma^2 t \). Using Chebyshev's inequality:
\[
P(|Y_t - \mathbb{E}[Y_t]| \geq \eta) \leq \frac{\sigma^2 T}{\eta^2},
\]
so by choosing a small enough \( \sigma \), \( Y_t \) concentrates around \( \mathbb{E}[Y_t] \).

In Appendix \ref{proof_conv}, we show that if samples are uniformly drawn from \( [0, T] \) and the number of samples increases, the spacing between adjacent samples converges to \( \frac{T}{N} \). Thus, by choosing appropriate inducing points \( \{(t_i, \vec{m}_i)\} \), as \( N \to \infty \), \( \mathbb{E}[Y_t] \) uniformly approximates any continuous function \( g(t) \) within any \( \varepsilon > 0 \), thereby establishing the universal approximation of our proposed model.

Here, we showed the universal approximation for 1-dimensional time series, highlighting the model's ability to capture the complex dynamics present in the time series. Therefore, we can expect that the model defined in Equations \ref{eq1} to \ref{eq6} to capture varying and diverse forms of temporal dynamics present in higher-dimensional time series.

\subsubsection{Avoiding Shrinkage of Inducing Points in Time}
We showed as the number of inducing points increases, the error in predicting the function approaches zero. However, in practice, we are interested in controlling the growth of inducing points and, consequently, the shrinkage of waiting times. This is motivated by our goal of obtaining a compact representation of the observed data, as a larger number of inducing points increases the computational cost of learning. While the independence assumption presented in section~\ref{hsde} simplifies model fitting, it introduces challenges such as the shrinkage of waiting times. To control the shrinkage, we set a prior on shape and scale of Gamma distribution that regulates both the expected waiting time and its variance. However, to better mitigate clustering and excessive shrinkage of waiting time, we can condition each waiting time on previous ones. One solution is to use a repulsive distribution \citep{ref35}, where the distribution of the current waiting time depends on previous ones. This is defined as:

\begin{equation}
p(\tau_1, \tau_2, \ldots, \tau_n) \propto \prod_{i=1}^{n} \mathrm{Gamma}(\tau_i; \beta_i, \cdots) \cdot \prod_{1 \leq i < j \leq n} \left(1 + \frac{\lambda}{|\tau_i - \tau_j|^2} \right)^{-1}
\end{equation}

Here, \(\lambda > 0\) controls the strength of the repulsion. This approach aligns with the history-based structure introduced in Equation \ref{eq1} and can be implemented by conditioning each \(\tau_i\) on a limited number of previous waiting times. Sampling from this prior is non-trivial, but it is feasible using importance sampling techniques\citep{ref47}.

\subsubsection{Nonparametric and Non-Markovian Properties}
In our model, the  $X_t$ exhibits dynamics similar to that of samples from a GP with specific covariance structures, for example, an Ornstein-Uhlenbeck process corresponding to an exponential kernel \citep{ref49}. Besides, the number of inducing points is not fixed in advance and it adapts dynamically to the complexity of the observed dynamics. The nonparametric and GP-like nature of our model makes it a alternative choice for machine learning and probabilistic applications. 

Trajectory of $X_t$ process is dependent to both past and future event-value pairs; thus it does not satisfy the Markovian property. This might complicate both training and inference within our framework. In section~\ref{hsde} , we reformulated the inducing point distribution using a renewal marked point process, which lets us to define $X_t$ and $Y_t$ process with a Markovian property. In section~\ref{inference}, we leverage this reformulation of SDEs for the inference and training of the model.

\subsubsection{Computational Cost}
A key advantage of the proposed model lies in its more favorable computational complexity compared to other non-parametric models such as GPs. Standard GPs require inversion of an $N \times N$ covariance matrix during their prediction, resulting in a computational complexity of $\mathcal{O}(N^3)$, where $N$ is the number of time points \citep{ref36}. This cubic scaling significantly limits the practicality of GPs in settings with long temporal sequences or high-frequency data. In contrast, as we see in the next section, inference for the discrete representation of our model can be performed using a sequential Monte Carlo (SMC) \citep{ref48} approach. When using particle-based methods such as Particle Marginal Metropolis-Hastings (PMMH), the computational cost per iteration scales as $\mathcal{O}(P \cdot N)$, where $P$ is the number of particles and $N$ again denotes the number of time points. This linear scaling with respect to $N$ enables the model to handle long trajectories more efficiently, making it well suited for high-resolution neural data. Moreover, since the model adaptively samples the trajectory space using a sparse set of inducing points, it provides even more scalable alternative to GP-based models without sacrificing expressive power.

\subsection{Model Training and Inference}
\label{inference}
The training objective is to maximize the marginal likelihood (or evidence) of the observed data $\{Z_k\}_{k=0}^K$. This involves updating several sets of parameters, including the event-value distribution parameters - i.e., the shape and scale parameters of the Gamma distribution, and the mean and covariance of the Gaussian distribution, the SDEs noise parameters ($\sigma_x^2$ and $\sigma_y^2$), and the observation model parameters ($W$ and $R$). Additionally, we must infer the trajectories of $X_t$ and $Y_t$ over $t = [0,T]$. For the training, we use an Expectation-Maximization (EM) algorithm, which can deal with latent process \citep{ref50}.

To develop the EM algorithm, we first derive a discrete-time representation of our model. To accomplish this, we use the renewal waiting time process discussed in section~\ref{hsde}. Under this assumption, we can break the non-Markovian dependence of $X_t$, which enables the use of recursive inference methods such as SMC. Simply put, at time $t$, we already know when the next inducing point occurs and its mark value, which breaks the dependence on the future trajectory of $X_t$. Appendix \ref{a33} discusses the discrete representation of SDEs. Given this representation, Algorithm 1 outlines the custom particle filtering algorithm developed for our model. A key component of this inference procedure is the proper sampling of event–value pairs, which is addressed within the algorithm. A more detailed explanation of this approach is provided in Appendix \ref{a23}.

\begin{algorithm}[h]
\label{algo}
\caption{SMC for Inducing Point and State Inference}
\begin{algorithmic}[1]
\State \textbf{Initialize:}
\State Set particles $U$, priors $p(x_0), p(y_0)$, proposal $\pi_k(x_k, y_k \mid x_{0:k-1}, y_{0:k-1}, z_k, \tau_{0:n_u}, m_{0:n_u})$, hyperparams $\alpha_0, \lambda_0, \mu_0, \xi_0$
\For{$u = 1$ to $U$}
    \State Sample $x_0^u \sim p(x_0)$, $y_0^u \sim p(y_0)$
    \State Set $m_0^u = \vec{0}, \tau_0^u = 0, n_u = 0, w_k^u = \frac{1}{U}$
\EndFor

\State \textbf{For each time step $k = 1$ to $K$:}
\For{$u = 1$ to $U$}
    \If{$k\Delta t > \tau_{\max(n_u)}^u$}
        \State Sample new $\tau^u \sim \Gamma(\alpha_0, \lambda_0)$, $m^u \sim \mathcal{N}(\mu_0, \xi_0)$
        \State Update particle with new $(\tau, m)$ and increment $n_u$
    \EndIf
    \State Sample $(x_k^u, y_k^u) \sim \pi_k(x_k, y_k \mid x_{0:k-1}^u, y_{0:k-1}^u, z_k, \tau_{0:n_u}^u, m_{0:n_u}^u)$
    \State Compute importance weight:
    \[
    w_k^u = w_{k-1}^u \cdot \frac{p(z_k \mid y_k^u) \cdot p(y_k^u \mid x_{k-1}^u) \cdot p(x_k^u \mid \tau_{0:n_u}^u, m_{0:n_u}^u)}{\pi_k(x_k^u, y_k^u \mid x_{0:k-1}^u, y_{0:k-1}^u, z_k, \tau_{0:n_u}^u, m_{0:n_u}^u)}
    \]
\EndFor

\State Normalize weights: $\hat{w}_k^u = \frac{w_k^u}{\sum_v w_k^v}$
\State Resample particles using $\hat{w}_k^u$, reset $w_k^u = \frac{1}{U}$
\end{algorithmic}
\label{alg1}
\end{algorithm}

In the M-step, we maximize the expectation of the complete log-likelihood with respect to the model parameters. Given the particle filters derived in the E-step, we find the MAP estimate for the shape and scale parameters of the waiting time model, as well as the mean and covariance of the mark distribution. We use a stochastic gradient ascent algorithm to optimize the observation model parameters \citep{ref51}. While it is possible to infer the stochastic process noise parameters, we generally set them manually to control the behavior of the SDEs processes. Further details of the optimization procedure are provided in Appendix~\ref{a43}.

Here, we defined both the training and inference procedures for one single observation of time series. In practice, however, this model can be applied to time series data collected across multiple trials of an experiment. In such cases, model parameters can be shared across trials, while each trial retains its own set of inducing points and corresponding $X_t$ and $Y_t$ trajectories.

\section{Results \& Discussion}
\subsection{Model Fit to 1-Dimensional Time Series Data}

\begin{wrapfigure}{r}{0.5\columnwidth}
    \centering
    \includegraphics[width=\linewidth]{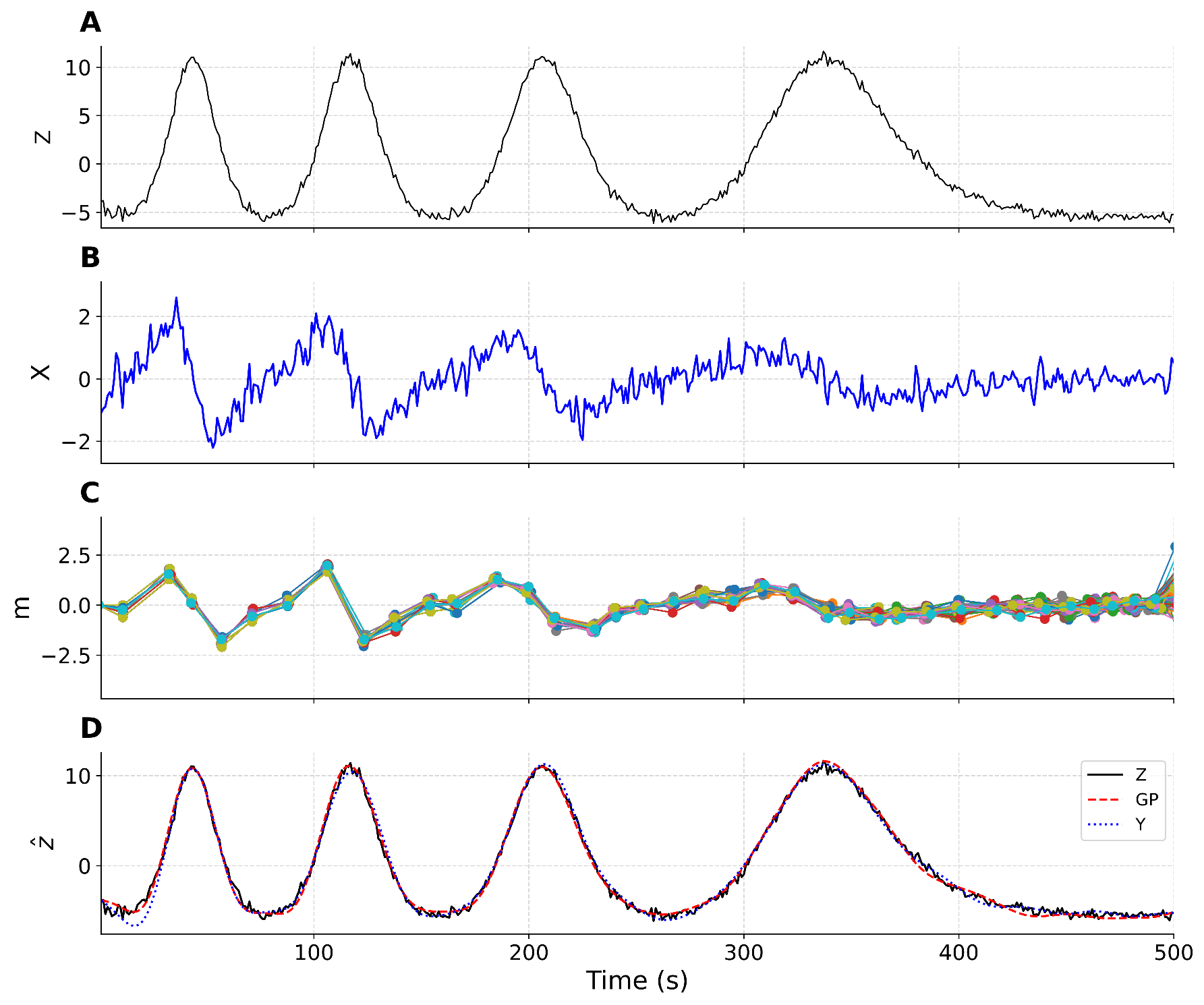}
    \caption{\textbf{Chirp Signal Reconstruction:} (A) Chirp signal with added white noise, (B) Posterior mean of $X$, (C) Mean values of $\tau$ and $m$ of the inducing points, (D) Prediction of the observed data using our framework and Gaussian Processes (GPs). The trajectory of $X$ closely follows the derivative of the chirp signal, and $Y$ accurately matches $Z$.}
    \label{fig:chirp}
\end{wrapfigure}

We applied our framework to a chirp signal lasting 250 seconds sampled at 2 Hz, resulting in 500 total samples. Additive white Gaussian noise ( with variance of 0.1) was added to each sample to produce (Figure~\ref{fig:chirp}A). For the model fit, the initial waiting times were sampled from a Gamma distribution parameterized to have a mean of 40 seconds and standard deviation of 8.94. We used the SMC with 10,000 particles with the SDEs noise parameters $\sigma_x$ and $\sigma_y$ set to \( 1 \times 10^{-1} \) and \( 1 \times 10^{-4} \) accordingly. For the model fit, we assume the variance of observation noise is known. Inference and parameter updates were run for 20 iterations.

Figure (2) shows the modeling results, which demonstrate that the model accurately reconstructs the chirp signal. During the inference process, the model dynamically adjusts the number of inducing points from an initial count of 13 to the final number of 29, with a waiting time of 9.09 in average. Given inducing points, we can generate the $X$ and then $Y$ trajectories for different time points. For the comparison, we used a GP model with a RBF kernel, where the GP gets values of $Z$ at 25 points in the time sampled uniformly with 10 seconds intervals. The mean squared error (MSE) of the prediction given GP over the whole signal and trajectory generated by our model are respectively 0.22 and 0.30 We expect that our model to have a slightly larger MSE; however, the critical advantage of our model is that it does not require kernel definition, its prediction of new points significantly requires less computation, and choice of inducing points is adaptive and is a part of model fit.

\subsection{Lorenz System Trajectory Decoder}

To illustrate the applicability of our method to decode underlying latent process through observing high-dimensional data, we applied it to a simulated data generated form the Lorenz attractor system \citep{ref42}. The original three-dimensional Lorenz dynamics were mapped into a 10-dimensional observation space via a random projection matrix, whose elements were drawn from a normal distribution. The observed signal consists of 500 points in time sampled at 10 Hz. Noise, sampled from a multivariate normal distribution, was added to the observations. The observation data, $Z$, is shown as a heatmap in Figure~\ref{fig:lorenz}B.

The initial parameters for the waiting-time distribution were consistent with those used in the chirp signal and the projection matrix was assumed to be known. Inference and parameter updates were carried out using a SMC algorithm with 20,000 particles over 20 iterations. Figure \ref{fig:lorenz}B illustrates both the simulated Lorenz attractor dynamics and the latent trajectories inferred by our model, with the inducing points marked along the trajectory. Our model’s inferred trajectory shows a close alignment with the simulated Lorenz system trajectory, demonstrating its capability to capture non-linear and volatile dynamics. It is clear more inducing points are placed around instances where there is a fast transition in the Lorenz dynamics suggesting model capability to adapt its waiting time according to the data attributes. It is worth to mention that inference of multi-dimensional state trajectory in GP requires proper definition of kernel and covariance structure to capture covariance across dimensions similarly we require an extensive data set for DKF and RNN based encoder-decoder models.

\subsection{Neural Population Dynamics During Reaching Tasks}

\begin{wrapfigure}{r}{0.5\columnwidth}
    \centering
    \includegraphics[width=\linewidth]{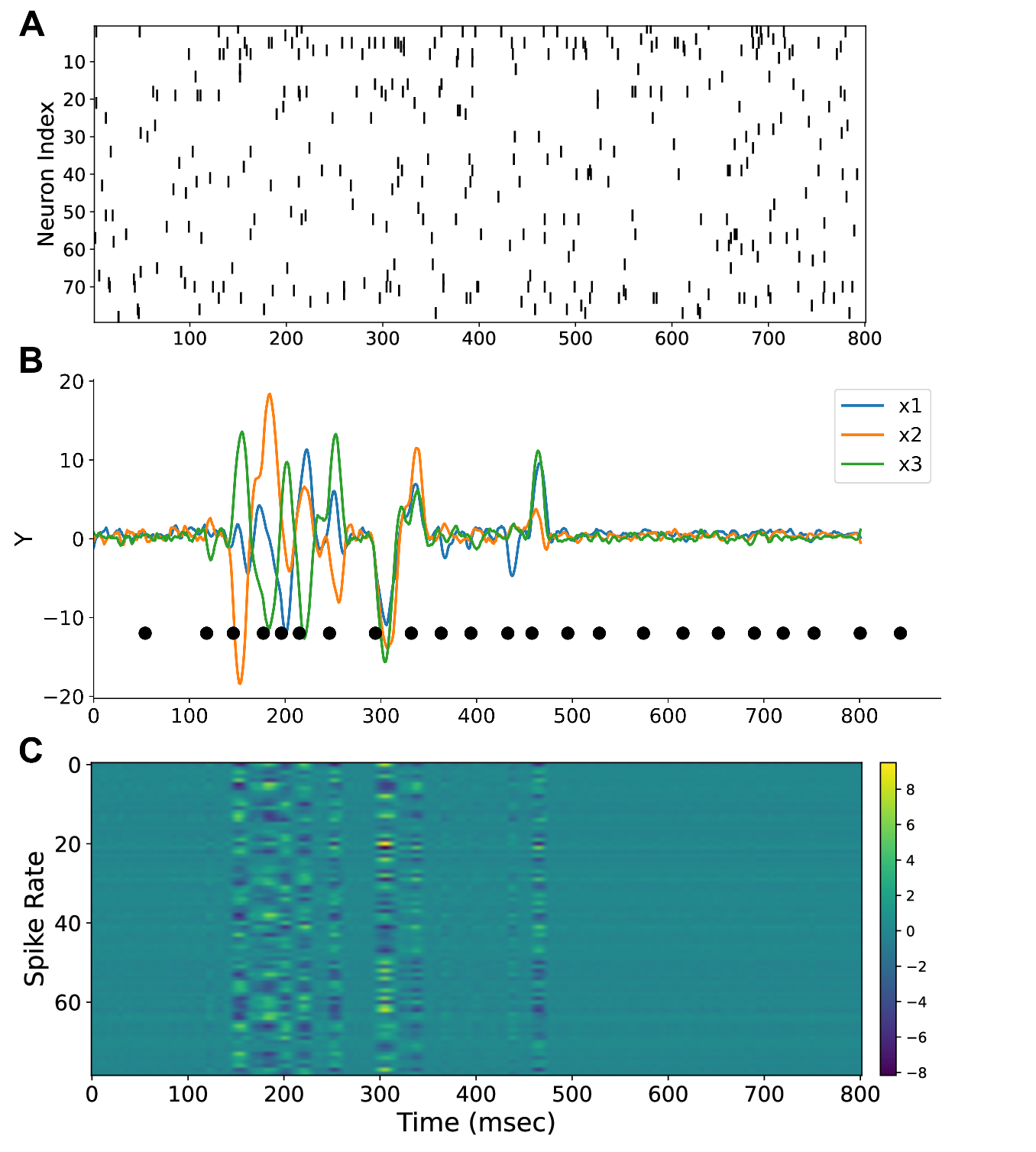}
    \caption{\textbf{Manifold Inference in Reach Task:} (A) Raster plot of neural activity, the target onset happens at 100 milliseconds. (B) Inferred manifold ($Y$) along with the timing of inducing points. (C) Instantaneous firing rate of a neuron. The inferred manifold along with instantaneous rate prediction suggest a significant change in neuronal ensemble shorty after target onset, reflecting diverse patterns of activity across neurons, as indicated by their estimated firing rates.}
    \label{fig:neural}
\end{wrapfigure}

In this example, we applied our framework to the Neural Latents Benchmark (NLB) MC\_Maze dataset \citep{ref43}. This dataset includes high-resolution electrophysiological recordings from macaque dorsal premotor (PMd) and primary motor (M1) cortices during a center-out reaching task \citep{ref40}. The dataset includes recording of 182 neurons activity sampled with 1 millisecond resolution as other covariates such as hand position, velocity and  cursor location are available as well.

Our modeling objective was to infer a low-dimensional representation of neural activity across different moments of the reach task. In particular, we focused on the period before and after target onset, as this window can reveal how neural activity forms a preparatory or cognitive map for reaching movements. We assumed the latent space comprising three state processes and observed spike data was characterized by a point process model \citep{ref52}. The choice of latent dimensionality is guided by model fit to the data, though it can be flexibly adjusted. We characterized each neuron's instantaneous firing rate as a linear combination of these latent states and assumed that neuronal activities are conditionally independent given the latent state process.

Out of 182 recorded neurons, we selected 79 that fired more than five spikes during the 3-second experiment. Figure 3 shows the inferred latent trajectories and the logarithm of each neuron's firing rate. It is worth noting that the weights for each neuron's rate model were tuned during model fitting (M-step). A noticeable change in the latent trajectory occurs shortly after target onset—starting around 100 ms post-onset and lasting approximately 400 ms. This change aligns with prior experimental findings, suggesting that neural activity encodes potential movements before their execution \citep{ref40}.

Here, we did not explore the relationship between the inferred latent states actual reach movements or task conditions, as it is outside the current scope. The model could also be reframed as a neural decoder, similar to our approach in the Lorenz benchmark problem; however, we did not apply that here, as this model attribute was discussed in a previous example.

\section{Discussion}
We developed the training and inference pipeline for a discretized version of our framework and applied it to simulated and neural datasets. The first example introduced the core concept of the model, showing its ability to draw proper inducing points that robustly captures temporal dynamics present in the data. The second example used high-dimensional simulated data with nonlinear and complex underlying dynamics, and highlights model capability to properly scale as the dimension of observations or latent states grow. In the third case, the model inference of latent processes aligns with the preparatory phase of motor task, despite there is no prior knowledge on how the neural activities encode movement. These collective results underscoring our proposed model usefulness in tasks like manifold discovery and neural decoding. Besides, these results corroborate with the universal approximation of the model as discussed in Section~\ref{secc}.

It is worth to mention that with other models such as DKF or GPs \citep{ref38} \citep{ref39} , we can infer the latent processes or reconstruct the data; however, they require large datasets, rely on strong priors – i.e., kernel or covariate choices, and involve more complex training and inference steps. Our model does not require a pre-defined kernel or extensive dataset and more importantly maintains a simple and interpretable structure. This is especially valuable in fields like neuroscience, where understanding the relationship between latent dynamics and observed data is crucial \citep{ref40}. Furthermore, our model offers dual interpretability, one where we can analyze the latent trajectories and their connection to other covariates, or we can examine the timing and values of the inducing points that shape the manifold.

We used a discretized representation of the model to derive its training and inference. However, extending to a continuous formulation — similar to neural ODEs \citep{ref41} — could improve inference robustness and will reduce numerical errors in generating states trajectory. Noise variance in both the $X$ and $Y$ processes strongly influences the model’s behavior. High variance in the drift complicates inducing points inference, while the variance in X controls the flexibility of the interpolation paths, from nearly linear to highly flexible. In current development of the model, we adjust their values by checking different values for both; however, tuning these parameters can be explored as part of the model training.

Training and inference in the model are performed using a custom SMC algorithm. While SMC in general can accommodate different observation types such as the point process observation model we used in neural data example, there may be more efficient algorithms for specific cases, such as the observation model defined in Equation (2). When event-value pairs are known, the model reduces to a linear state-space system, allowing use of the Kalman filter, which provides faster and more robust estimation. This motivates a hybrid approach, which we can use SMC for the event-value inference and the Kalman filter for state estimation, which will be explored in further development of this framework.

Currently, we assume independence between marks and waiting times, and no dependence on past events. This simplifies the model inference and training but may reduce its ability to adjust inducing points to capture changes in data happening at different temporal scales. For example, fast transition in the latent state may require shorter waiting times and larger marks. As a result, exploring alternative distributions for mark and waiting time—including those that are time - or history - dependent—may be necessary to improve model performance.

Lastly, we think conducting an extensive comparative analysis with existing methods to better highlight the distinct advantages and limitations of our model are needed. This will be another aspect of the future work.

\section{Conclusion}
We introduced a hierarchical SDE framework that uncovers low-dimensional manifolds in high-dimensional time-series data by inferring a sparse set of inducing points across time and value dimensions. The model supports efficient inference and learning with linear computational complexity in both data dimensions and time. Its flexibility and expressiveness make it well suited for neural data with complex dynamics, and it holds promise as an unsupervised dimensionality reduction tool, dynamical neural decoder, or adaptive feature extractor for a range of neuroscience applications.

\bibliographystyle{abbrvnat}
\bibliography{ref}

\begin{thebibliography}{39}
\providecommand{\natexlab}[1]{#1}
\providecommand{\url}[1]{\texttt{#1}}
\expandafter\ifx\csname urlstyle\endcsname\relax
  \providecommand{\doi}[1]{doi: #1}\else
  \providecommand{\doi}{doi: \begingroup \urlstyle{rm}\Url}\fi

\bibitem[Bondanelli et~al.(2021)Bondanelli, Deneux, Bathellier, and Ostojic]{ref19}
G.~Bondanelli, T.~Deneux, B.~Bathellier, and S.~Ostojic.
\newblock Network dynamics underlying off responses in the auditory cortex.
\newblock \emph{Elife}, 10:\penalty0 e53151, 2021.

\bibitem[Brown and Kass(2018)]{ref50}
E.~N. Brown and R.~E. Kass.
\newblock Estimating a state-space model from point process observations.
\newblock \emph{Unpublished Manuscript}, 2018.

\bibitem[Casale et~al.(2018)Casale, Dalca, Saglietti, Listgarten, and Fusi]{ref39}
F.~P. Casale, A.~V. Dalca, L.~Saglietti, J.~Listgarten, and N.~Fusi.
\newblock Gaussian process prior variational autoencoders.
\newblock 31, 2018.

\bibitem[Chang et~al.(2024)Chang, Zhang, Wang, Qin, Zhao, and Wang]{ref11}
H.~Chang, Q.~Zhang, Y.~Wang, Z.~Qin, L.~Zhao, and H.~Wang.
\newblock Unlocking the power of lstm for long term time series forecasting.
\newblock \emph{Proceedings of the AAAI Conference on Artificial Intelligence}, 38\penalty0 (4):\penalty0 4292--4300, 2024.

\bibitem[Chen et~al.(2018)Chen, Rubanova, Bettencourt, and Duvenaud]{ref41}
R.~T.~Q. Chen, Y.~Rubanova, J.~Bettencourt, and D.~Duvenaud.
\newblock Neural ordinary differential equations.
\newblock 31, 2018.

\bibitem[Churchland et~al.(2012)Churchland, Cunningham, Kaufman, Ryu, and Shenoy]{ref40}
M.~M. Churchland, J.~P. Cunningham, M.~T. Kaufman, S.~I. Ryu, and K.~V. Shenoy.
\newblock Neural population dynamics during reaching.
\newblock \emph{Nature}, 487\penalty0 (7405):\penalty0 51--56, 2012.

\bibitem[Cunningham and Yu(2014)]{ref34}
J.~P. Cunningham and B.~M. Yu.
\newblock Dimensionality reduction for large-scale neural recordings.
\newblock \emph{Nature neuroscience}, 17\penalty0 (11):\penalty0 1500--1509, 2014.

\bibitem[Daley and Vere-Jones(2006)]{ref15}
D.~J. Daley and D.~Vere-Jones.
\newblock \emph{An introduction to the theory of point processes: volume I: elementary theory and methods}.
\newblock Springer Science \& Business Media, 2006.

\bibitem[Doucet et~al.(2001)Doucet, de~Freitas, and Gordon]{ref48}
A.~Doucet, N.~de~Freitas, and N.~Gordon.
\newblock \emph{Sequential Monte Carlo Methods in Practice}.
\newblock Springer, New York, 2001.

\bibitem[Eden et~al.(2004)Eden, Frank, Barbieri, Solo, and Brown]{ref52}
U.~T. Eden, L.~M. Frank, R.~Barbieri, V.~Solo, and E.~N. Brown.
\newblock Dynamic analysis of neural encoding by point process adaptive filtering.
\newblock \emph{Neural Computation}, 16\penalty0 (5):\penalty0 971--998, 2004.

\bibitem[Eleftheriadis et~al.(2017)Eleftheriadis, Nicholson, Deisenroth, and Hensman]{ref5}
S.~Eleftheriadis, T.~Nicholson, M.~Deisenroth, and J.~Hensman.
\newblock Identification of gaussian process state space models.
\newblock \emph{Advances in neural information processing systems}, 30, 2017.

\bibitem[Fox et~al.(2008)Fox, Sudderth, Jordan, and Willsky]{ref24}
E.~Fox, E.~Sudderth, M.~Jordan, and A.~Willsky.
\newblock Nonparametric bayesian learning of switching linear dynamical systems.
\newblock In D.~Koller, D.~Schuurmans, Y.~Bengio, and L.~Bottou, editors, \emph{Advances in Neural Information Processing Systems}, volume~21. Curran Associates, Inc., 2008.

\bibitem[Ghahramani and Hinton(2000)]{ref23}
Z.~Ghahramani and G.~E. Hinton.
\newblock Variational learning for switching state-space models.
\newblock \emph{Neural Computation}, 12\penalty0 (4):\penalty0 831--864, 2000.
\newblock \doi{10.1162/089976600300015619}.

\bibitem[Girin et~al.(2020)Girin, Leglaive, Bie, Diard, Hueber, and Alameda-Pineda]{ref22}
L.~Girin, S.~Leglaive, X.~Bie, J.~Diard, T.~Hueber, and X.~Alameda-Pineda.
\newblock Dynamical variational autoencoders: A comprehensive review.
\newblock \emph{arXiv preprint arXiv:2008.12595}, 2020.

\bibitem[Glorot and Bengio(2010)]{ref10}
X.~Glorot and Y.~Bengio.
\newblock Understanding the difficulty of training deep feedforward neural networks.
\newblock In \emph{Proceedings of the Thirteenth International Conference on Artificial Intelligence and Statistics (AISTATS)}, pages 249--256. PMLR, 2010.

\bibitem[Gonzalez-Castillo et~al.(2023)Gonzalez-Castillo, Fernandez, Lam, Handwerker, Pereira, and Bandettini]{ref20}
J.~Gonzalez-Castillo, I.~S. Fernandez, K.~C. Lam, D.~A. Handwerker, F.~Pereira, and P.~A. Bandettini.
\newblock Manifold learning for fmri time-varying functional connectivity.
\newblock \emph{Frontiers in Human Neuroscience}, 17:\penalty0 1134012, 2023.

\bibitem[Gosztolai et~al.(2023)Gosztolai, Peach, Arnaudon, Barahona, and Vandergheynst]{ref55}
A.~Gosztolai, R.~L. Peach, A.~Arnaudon, M.~Barahona, and P.~Vandergheynst.
\newblock Interpretable statistical representations of neural population dynamics and geometry.
\newblock \emph{arXiv preprint}, 2023.

\bibitem[Hastings(1970)]{ref47}
W.~K. Hastings.
\newblock Monte carlo sampling methods using markov chains and their applications.
\newblock \emph{Biometrika}, 57\penalty0 (1):\penalty0 97--109, 1970.

\bibitem[Jacobsen(2006)]{ref44}
M.~Jacobsen.
\newblock \emph{Point Process Theory and Applications: Marked Point and Piecewise Deterministic Processes}.
\newblock Birkhäuser, Boston, 2006.

\bibitem[Krishnan et~al.(2015{\natexlab{a}})Krishnan, Shalit, and Sontag]{ref3}
R.~G. Krishnan, U.~Shalit, and D.~Sontag.
\newblock Deep kalman filters, 2015{\natexlab{a}}.

\bibitem[Krishnan et~al.(2015{\natexlab{b}})Krishnan, Shalit, and Sontag]{ref38}
R.~G. Krishnan, U.~Shalit, and D.~Sontag.
\newblock Deep kalman filters.
\newblock \emph{arXiv preprint arXiv:1511.05121}, 2015{\natexlab{b}}.

\bibitem[Kwon et~al.(2020)Kwon, Oh, and Lim]{ref21}
J.~Kwon, H.-S. Oh, and Y.~Lim.
\newblock Dynamic principal component analysis with missing values.
\newblock \emph{Journal of Applied Statistics}, 47\penalty0 (11):\penalty0 1957--1969, 2020.

\bibitem[Lorenz(1963)]{ref42}
E.~N. Lorenz.
\newblock Deterministic nonperiodic flow.
\newblock \emph{Journal of the Atmospheric Sciences}, 20\penalty0 (2):\penalty0 130--141, 1963.

\bibitem[McInnes et~al.(2018)McInnes, Healy, and Melville]{ref31}
L.~McInnes, J.~Healy, and J.~Melville.
\newblock Umap: Uniform manifold approximation and projection for dimension reduction.
\newblock \emph{arXiv preprint arXiv:1802.03426}, 2018.

\bibitem[Oksendal(2013)]{ref16}
B.~Oksendal.
\newblock \emph{Stochastic differential equations: an introduction with applications}.
\newblock Springer Science \& Business Media, 2013.

\bibitem[Pei et~al.(2021)Pei, Ye, Zoltowski, Wu, Chowdhury, Sohn, O'Doherty, Shenoy, Kaufman, Churchland, Jazayeri, Miller, Pillow, Park, Dyer, and Pandarinath]{ref43}
F.~Pei, J.~Ye, D.~M. Zoltowski, A.~Wu, R.~H. Chowdhury, H.~Sohn, J.~E. O'Doherty, K.~V. Shenoy, M.~T. Kaufman, M.~M. Churchland, M.~Jazayeri, L.~E. Miller, J.~W. Pillow, I.~M. Park, E.~L. Dyer, and C.~Pandarinath.
\newblock Neural latents benchmark '21: Evaluating latent variable models of neural population activity.
\newblock \emph{Advances in Neural Information Processing Systems}, 2021.

\bibitem[Petralia et~al.(2012)Petralia, Rao, and Dunson]{ref35}
F.~Petralia, V.~Rao, and D.~Dunson.
\newblock Repulsive mixtures.
\newblock \emph{Advances in neural information processing systems}, 25, 2012.

\bibitem[Pitman and Yor(1999)]{ref45}
J.~Pitman and M.~Yor.
\newblock Brownian bridge and related stochastic processes.
\newblock \emph{Probability Surveys}, 1:\penalty0 1--61, 1999.

\bibitem[Rangapuram et~al.(2018)Rangapuram, Seeger, Gasthaus, Stella, Wang, and Januschowski]{ref2}
S.~S. Rangapuram, M.~W. Seeger, J.~Gasthaus, L.~Stella, Y.~Wang, and T.~Januschowski.
\newblock Deep state space models for time series forecasting.
\newblock \emph{Advances in neural information processing systems}, 31, 2018.

\bibitem[Rusch and Rus(2025)]{ref54}
T.~K. Rusch and D.~Rus.
\newblock Linear oscillatory state-space models.
\newblock \emph{Proceedings of the International Conference on Learning Representations (ICLR)}, 2025.

\bibitem[S{\"a}rkk{\"a} and Svensson(2023)]{ref1}
S.~S{\"a}rkk{\"a} and L.~Svensson.
\newblock \emph{Bayesian filtering and smoothing}, volume~17.
\newblock Cambridge university press, 2023.

\bibitem[Seeger(2004)]{ref36}
M.~Seeger.
\newblock Gaussian processes for machine learning.
\newblock \emph{International journal of neural systems}, 14\penalty0 (02):\penalty0 69--106, 2004.

\bibitem[Shannon(1949)]{ref33}
C.~E. Shannon.
\newblock Communication in the presence of noise.
\newblock \emph{Proceedings of the IRE}, 37\penalty0 (1):\penalty0 10--21, 1949.

\bibitem[Uhlenbeck and Ornstein(1930)]{ref49}
G.~E. Uhlenbeck and L.~S. Ornstein.
\newblock On the theory of the brownian motion.
\newblock \emph{Physical Review}, 36\penalty0 (5):\penalty0 823--841, 1930.

\bibitem[Van~der Maaten and Hinton(2008)]{ref30}
L.~Van~der Maaten and G.~Hinton.
\newblock Visualizing data using t-sne.
\newblock \emph{Journal of machine learning research}, 9\penalty0 (11), 2008.

\bibitem[Wang et~al.(2005)Wang, Hertzmann, and Fleet]{ref4}
J.~Wang, A.~Hertzmann, and D.~J. Fleet.
\newblock Gaussian process dynamical models.
\newblock \emph{Advances in neural information processing systems}, 18, 2005.

\bibitem[Whiteley et~al.(2024)Whiteley, Gray, and Rubin-Delanchy]{ref18}
N.~Whiteley, A.~Gray, and P.~Rubin-Delanchy.
\newblock Statistical exploration of the manifold hypothesis, 2024.

\bibitem[Xu et~al.(2021)Xu, Zhang, and Lu]{ref51}
Z.~Xu, L.~Zhang, and Y.~Lu.
\newblock Variational posterior approximation using stochastic gradient ascent for dirichlet process mixture models.
\newblock \emph{Pattern Recognition}, 110:\penalty0 107637, 2021.

\bibitem[Ziaei et~al.(2024)Ziaei, Nazari, Eden, Widge, and Yousefi]{ref6}
N.~Ziaei, B.~Nazari, U.~T. Eden, A.~Widge, and A.~Yousefi.
\newblock A bayesian gaussian process-based latent discriminative generative decoder (ldgd) model for high-dimensional data.
\newblock \emph{IEEE Access}, 2024.

\end{thebibliography}

\appendix
\section{Appendix}

\subsection{Gamma Distribution for Waiting Times and Prior Selection for Inducing Points}
\label{app:priors}

To complete the Bayesian framework, we define priors for the model parameters. For the mark distribution parameters, we assume:
\begin{equation}
    \mu_i \mid \Sigma_i \sim N(\mu_0, \lambda \Sigma_i), \quad \Sigma_i \sim \text{Inverse-Wishart}(\nu, \Psi)
\end{equation}

where \( \mu_0 \), \( \lambda \), \( \nu \), and \( \Psi \) are hyperparameters. For the Gamma distribution parameters \( \alpha \) and \( \lambda \) governing the waiting times \( \tau_i \), we consider the following prior options based on domain-specific knowledge, though their specific forms remain to be fully specified in this study:
\begin{itemize}
    \item For \( \alpha \):
    \begin{itemize}
        \item \( \alpha \sim \text{Gamma}(a_0, b_0) = \frac{b_0^{a_0}}{\Gamma(a_0)} \alpha^{a_0 - 1} e^{-b_0 \alpha} \),
        \item \( \alpha \sim \text{Exp}(\lambda_0) = \lambda_0 e^{-\lambda_0 \alpha} \),
        \item \( \alpha \sim \text{Lognormal}(\mu_0, \sigma_0^2) = \frac{1}{\alpha \sqrt{2\pi \sigma_0^2}} \exp\left(-\frac{(\log \alpha - \mu_0)^2}{2\sigma_0^2}\right) \),
    \end{itemize}
    \item For \( \lambda \):
    \begin{itemize}
        \item \( \lambda \sim \text{Gamma}(c_0, d_0) = \frac{d_0^{c_0}}{\Gamma(c_0)} \lambda^{c_0 - 1} e^{-d_0 \lambda} \),
        \item \( \lambda \sim \text{InvGamma}(\gamma_0, \delta_0) = \frac{\delta_0^{\gamma_0}}{\Gamma(\gamma_0)} \lambda^{-(\gamma_0 + 1)} e^{-\delta_0 / \lambda} \),
    \end{itemize}
\end{itemize}
where \( a_0, b_0, \lambda_0, \mu_0, \sigma_0^2, c_0, d_0, \gamma_0, \delta_0 \) are hyperparameters.

In our model, the waiting times \(\tau_i\) and the marks \(\vec{m}_i\) associated with each event are generated according to specific probabilistic distributions:

\begin{itemize}
\item \textbf{Waiting Times \(\tau_i\):} \\
The waiting times between events are assumed to follow a Gamma distribution parameterized by a shape parameter \(\alpha\) and a rate parameter \(\lambda\). The probability density function for \(\tau_i\) is given by:
\begin{equation}
p(\tau_i) = \text{Gamma}(\tau_i; \alpha, \lambda)
\end{equation}
where the Gamma distribution is defined as:
\begin{equation}
\text{Gamma}(\tau_i; \alpha, \lambda) = \frac{\lambda^\alpha}{\Gamma(\alpha)} \tau_i^{\alpha - 1} e^{-\lambda \tau_i}, \quad \tau_i > 0
\end{equation}
and \(\Gamma(\alpha)\) denotes the Gamma function evaluated at \(\alpha\).

\medskip

\textbf{Motivation for using the Gamma distribution:}

Consider \( N \) i.i.d.\ samples \( U_1, \ldots, U_N \sim \text{Uniform}(0, T) \), and denote their order statistics by \( U_{(1)} \le \cdots \le U_{(N)} \). Define the gaps between consecutive order statistics as
\begin{equation}
\Delta_0 = U_{(1)}, \quad \Delta_i = U_{(i+1)} - U_{(i)} \text{ for } i = 1, \ldots, N-1, \quad \Delta_N = T - U_{(N)} \tag{14}
\end{equation}

As \( N \to \infty \), it is well-known that each gap satisfies \( \Delta_i \xrightarrow{p} T/N \), and the rescaled gaps \( N \Delta_i \) converge in distribution to an exponential random variable, that is,
\begin{equation}
N \Delta_i \xrightarrow{d} \text{Exp}(1) \tag{15}
\end{equation}

Moreover, the normalized gaps \( (\Delta_0/T, \ldots, \Delta_N/T) \) jointly follow a Dirichlet\( (1, \ldots, 1) \) distribution. Marginally, each normalized gap \( \Delta_i / T \) follows a Beta\( (1, N) \) distribution. As \( N \) becomes large, the Beta\( (1, N) \) distribution approximates a Gamma\( (1, 1/N) \) distribution, because
\begin{equation}
N \cdot (\Delta_i / T) \xrightarrow{d} \text{Exp}(1) \tag{16}
\end{equation}

which suggests that
\begin{equation}
\Delta_i \approx \text{Gamma}(1, T/N) \tag{17}
\end{equation}

Thus, in the large-sample limit, the gaps between ordered uniform samples behave approximately like scaled exponential random variables.

To simulate ordered points efficiently for a finite number \( M \) of samples, we propose sampling \( M \) independent gaps
\begin{equation}
\Delta_i \sim \text{Gamma}(1, T/M) \tag{18}
\end{equation}
and constructing ordered points via the cumulative sums
\begin{equation}
U_{(i)} = \sum_{j=0}^{i-1} \Delta_j, \quad i = 1, \ldots, M \tag{19}
\end{equation}

This motivates our use of Gamma-distributed waiting times \(\tau_i\) in the model, capturing the natural variability in the timing of events.

\item \textbf{Marks \(\vec{m}_i\):} \\
The marks, representing additional information associated with each event, are modeled as drawn from a multivariate normal (Gaussian) distribution. Each mark vector \(\vec{m}_i\) has an associated mean vector \(\mu_i\) and covariance matrix \(\Sigma_i\), with the distribution:
\begin{equation}
p(\vec{m}_i) = \mathcal{N}(\vec{m}_i; \mu_i, \Sigma_i)
\end{equation}
explicitly given by:
\begin{equation}
\mathcal{N}(\vec{m}_i; \mu_i, \Sigma_i) = \frac{1}{(2\pi)^{d/2} |\Sigma_i|^{1/2}} \exp\left(-\frac{1}{2} (\vec{m}_i - \mu_i)^\top \Sigma_i^{-1} (\vec{m}_i - \mu_i)\right)
\end{equation}
where \(d\) is the dimensionality of the mark vector.
\end{itemize}

This modeling choice allows flexible and realistic characterization of the temporal dynamics \(\tau_i\) and the event-related features \(\vec{m}_i\) within the system under study.

\subsection{Convergence of Spacing Between Adjacent Samples}
\label{proof_conv}

Let \( X_1, X_2, \ldots, X_N \) be i.i.d. random variables uniformly distributed on \([0, T]\), and let \( X_{(1)} \leq X_{(2)} \leq \cdots \leq X_{(N)} \) denote their order statistics. Define the adjacent spacings \( d_i = X_{(i+1)} - X_{(i)} \) for \( i = 1, \ldots, N - 1 \).

From properties of uniform order statistics, the expected value of the \( i \)-th order statistic is \( \mathbb{E}[X_{(i)}] = \frac{iT}{N+1} \). It follows that

\begin{equation}
    \mathbb{E}[d_i] = \mathbb{E}[X_{(i+1)} - X_{(i)}] = \frac{T}{N+1}
\end{equation}

which satisfies \( \mathbb{E}[d_i] \to \frac{T}{N} \) as \( N \to \infty \).

Moreover, the variance of \( d_i \) satisfies \( \mathrm{Var}(d_i) = O(N^{-2}) \). By Chebyshev’s inequality,

\begin{equation}
    \mathbb{P}\left( \left| d_i - \mathbb{E}[d_i] \right| \geq \epsilon \right) \leq \frac{\mathrm{Var}(d_i)}{\epsilon^2} = O\left( \frac{1}{N^2} \right)
\end{equation}

which vanishes as \( N \to \infty \). Therefore, \( d_i \to \frac{T}{N} \) in probability.

Furthermore, classical results on uniform spacings imply that the maximum spacing \( \max_i d_i \) satisfies

\begin{equation}
    \max_{1 \leq i \leq N-1} d_i = \frac{T}{N} + O\left(N^{-1/2}\right)
\end{equation}

with high probability, confirming that all gaps become uniformly close to \( \frac{T}{N} \) as \( N \to \infty \).

\subsection{Discrete Representation of Hierarchical SDE}
\label{a33}

We focus on a discrete-time formulation of the model. To construct the discrete process, we assume that \( X_t \) and \( Y_t \) are sampled at regular intervals of \( \Delta t \). The discrete representation of \( x_t^d \) is defined as:

\begin{equation}
\begin{split}
x_{k+1}^d =\ & x_k^d + \frac{m_{i+1}^d - x_k^d}{t_{i+1} - k \Delta t} \, \Delta t + \sqrt{\frac{(t_{i+1} - k \Delta t)(k \Delta t - t_i)}{t_{i+1} - t_i}\Delta t} \, \cdot w_k^d, \\
& w_k^d \sim \mathcal{N}(0, \sigma_x^{d^2})
\end{split}
\end{equation}

where \( w_k^d \) is a Gaussian noise term. Similarly, the discrete-time evolution of \( y_t^d \) is:

\begin{equation}
y_{k+1}^d = y_k^d + x_k^d \, \Delta t + \sqrt{\Delta t} \cdot \nu_k^d, \quad \nu_k^d \sim \mathcal{N}(0, \sigma_y^{d^2})
\end{equation}

where \( \nu_k^d \) is Gaussian noise. The discrete observation process is given by:

\begin{equation}
Z_k = W Y_k + \xi_k, \quad \xi_k \sim \mathcal{N}(0, R)
\end{equation}

where \( Y_k = \begin{pmatrix} y_k^1 \\ \vdots \\ y_k^D \end{pmatrix} \), \( W \in \mathbb{R}^{M \times D} \) is a projection matrix, and \( \xi_k \) is observation noise.

To ensure accuracy, \( \Delta t \) must be much smaller than the minimum inter-event time, i.e., \( \Delta t \ll \min(\tau_i) \), so that no two inducing points fall within the same discrete time bin. This constraint can be satisfied by analyzing the posterior distribution of waiting times and adjusting \( \Delta t \) accordingly. In essence, we require an orderly event process---allowing at most one event per bin---which can be enforced by carefully selecting the bin size \( \Delta t \).

When using the waiting time representation, at time $k$, we already know when the next inducing point (e.g., event 213) will occur and what its value will be. From a Markovian perspective, we can assume that the entire process is determined at time $k$. This assumption simplifies the inference procedure presented in Algorithm~\ref{alg1}.

\subsection{Detailed Version of the SMC Algorithm}
\label{a23}
Here, we provide a more detailed description of the SMC algorithm introduced in the main text for inference in our model. This is presented in Algorithm 2.

\begin{algorithm}[t]
\caption{SMC Algorithm For Inferring Inducing Points and State Estimation}
\begin{algorithmic}[1]
\State \textbf{Set Algorithm Hyperparameters:}
\State Set number of particles $U$
\State Define initial distributions $p(x_0)$ and $p(y_0)$
\State Define proposal density $\pi_k(x_k, y_k \mid x_{0:k-1}, y_{0:k-1}, z_k, \tau_{0:n_u}, m_{0:n_u})$
\State Set hyperparameters $\alpha_0, \lambda_0$ For $\tau$ distribution
\State Set hyperparameters $\mu_0, \xi_0$ For $m$ distribution

\vspace{1mm}
\State \textbf{Initialization:}
\For{$u = 1$ to $U$}
    \State Sample $x_0^u \sim p(x_0)$, $y_0^u \sim p(y_0)$
    \State Set $m_0^u = \vec{0}$, $\tau_0^u = 0$, $n_u = 0$
    \State Set initial weight $w_k^u = \frac{1}{U}$
    \State Initialize particle $D_0^u = \{x_0^u, y_0^u, \tau_0^u, m_0^u, n_u\}$
\EndFor

\vspace{1mm}
\State \textbf{Inference:}
\For{$k = 1$ to $K$}

    \State \textbf{1. Time \& Mark Sampling:}
    \For{$u = 1$ to $U$}
        \If{$k \cdot \Delta t > \tau_{\max(n_u)}^u$}
            \State Sample $\tau_{\text{new}}^u \sim \Gamma(\tau; \alpha_0, \lambda_0)$
            \State Sample $m_{\text{new}}^u \sim \mathcal{N}(m; \mu_0, \xi_0)$
            \State Update $D_k^u = \{x_{0:k-1}^u, y_{0:k-1}^u, \tau_{0:n_u}^u, m_{0:n_u}^u, \tau_{\text{new}}^u, m_{\text{new}}^u, n_u+1\}$
        \EndIf
    \EndFor

    \State \textbf{2. Sampling:}
    \For{$u = 1$ to $U$}
        \State Sample $(x_k^u, y_k^u) \sim \pi_k(x_k, y_k \mid x_{0:k-1}^u, y_{0:k-1}^u, z_k, \tau_{0:n_u}^u, m_{0:n_u}^u)$
        \State Compute importance weight:
        \[
        w_k^u = w_{k-1}^u \cdot \frac{p(z_k \mid y_k^u) \cdot p(y_k^u \mid x_{k-1}^u) \cdot p(x_k^u \mid \tau_{0:n_u}^u, m_{0:n_u}^u)}{\pi_k(x_k^u, y_k^u \mid x_{0:k-1}^u, y_{0:k-1}^u, z_k, \tau_{0:n_u}^u, m_{0:n_u}^u)}
        \]
    \EndFor

    \State \textbf{3. Normalization:}
    \For{$u = 1$ to $U$}
         \[ \hat{w}_k^u = \frac{w_k^u}{\sum_{v=1}^{U} w_k^v}\]
    \EndFor

    \State \textbf{4. Resampling:}
    \State Resample $U$ particles $D_k^u = \{x_{0:k}^u, y_{0:k}^u, \tau_{0:n_u}^u, m_{0:n_u}^u, n_u\}$ from $\{D_k^u\}_{u=1}^{U}$ with probabilities $\hat{w}_k^u$
    \For{$u = 1$ to $U$}
        \State Reset weight: $w_k^u = \frac{1}{U}$
    \EndFor

\EndFor
\end{algorithmic}
\label{alg2}
\end{algorithm}

\subsection{Training Step: M-step}
\label{a43}
For the M-step, we assume that the SMC algorithm has been run and that we have obtained \( D_K^u \) for \( u = 1, \ldots, U \). The full likelihood of the process is defined as:

\begin{align}
\label{eq17}
P(Z_{1:K}, X_{0:K}, Y_{0:K}, \tau_{1:n_u}, \vec{m}_{1:n_u}; \omega, \omega_0)
&= P(X_0, Y_0) \prod_{k=1}^{K} p(Z_k \mid Y_k, W, R) p(Y_k \mid X_{k-1}, Y_{k-1}, \sigma_y) \notag \\
&\quad \times p(X_k \mid X_{k-1}, \tau_{1:n_s}, \sigma_y, \vec{m}_{1:n_s}) \prod_{n=1}^{n_u} p(\tau_n) p(\vec{m}_n) p(\omega_0^n)
\end{align}

Here, \( \omega \) represents the model parameters \( \{W, R, \sigma_y, \sigma_x\} \), and \( \omega_0 \) is the set of hyperparameters defining the priors, as detailed in Appendix A.1. The term \( p(\omega_0^n) \) appears for each evnt-mark pair because the prior is applied individually to each waiting time and mark.

In the M-step, we compute the expectation of the full log-likelihood with respect to the posterior distribution over the latent processes and variables in the model. The latent processes are denoted by \( X \) and \( Y \), while \( t_i \) and \( m_i \) represent another set of latent variables. The \( Q \)-function, with respect to which the expectation is taken, is defined as:

\begin{align}
Q &= \mathbb{E}_{p(X_{0:K}, Y_{0:K}, \tau_{1:n_u}, \vec{m}_{1:n_u} \mid Z_{1:K}, \omega)}
\left[ \log P(Z_{1:K}, X_{0:K}, Y_{0:K}, \tau_{1:n_u}, \vec{m}_{1:n_u}; \omega) \right] \notag \\
&= \sum_{u=1}^U \log P(X_0^u, Y_0^u) +
\sum_{u=1}^U \sum_{k=1}^K \log p(Z_k \mid Y_k^u, W, R) +
\sum_{u=1}^U \sum_{k=1}^K \log p(Y_k^u \mid X_{k-1}^u, Y_{k-1}^u, \sigma_x) \notag \\
&\quad + \sum_{u=1}^U \sum_{k=1}^K \log p(X_k^u \mid X_{k-1}^u, \tau_{1:n_s}^u, \sigma_y, \vec{m}_{1:n_s}^u; n_s = \min_n \tau_n^u > k) \notag \\
&\quad + \sum_{u=1}^U \sum_{n=1}^{n_u} \log p(\tau_n^u) +
\sum_{u=1}^U \sum_{n=1}^{n_u} \log p(\vec{m}_n^u) +
\sum_{n=1}^{n_u} \log p(\omega_0^n)
\end{align}

For the observation model defined in Equation~\ref{eq6}, the estimation of parameters \( W \) and \( R \) corresponds to a multivariate linear regression fit to samples of the \( X \) trajectory. Thus, \( W \) and \( R \) can be estimated in closed form, similar to the approach used in linear regression. The waiting time distribution parameters, i.e., the shape and scale, and the mark distribution parameters, i.e., mean and covariance, are estimated via MAP using optimization routines.

Given the model formulation, we require running \( 2n_u \) (maximum number of inducing points generated by the SMC algorithm): one per waiting time and one per mark. Although both \( \sigma_y \) and \( \sigma_x \) can be learned, we typically fix \( \sigma_y \) to ensure meaningful propagation from \( Y \) to \( X \). If \(\sigma_y\) is too large, changes in \( Z \) are mostly captured by shifting \( Y \), which limits the propagation of observed data information to \( X \) and the inducing points. On the other hand, \(\sigma_x\) can be optimized, and a closed-form solution for its estimation can be derived. Similar to the waiting time and mark parameters, we can use optimization techniques for its estimation. 

In Equation~\ref{eq17}, we assume shared \( \sigma_y \) and \( \sigma_x \) across latent dimensions, in practice, these can vary per dimension.

\newpage

\end{document}